\documentclass{article}

\usepackage{PRIMEarxiv}

\usepackage[utf8]{inputenc} 
\usepackage[T1]{fontenc}    
\usepackage{hyperref}       
\usepackage{url}            
\usepackage{booktabs}       
\usepackage{amsfonts}       
\usepackage{nicefrac}       
\usepackage{microtype}      
\usepackage{lipsum}
\usepackage{fancyhdr}       
\usepackage{graphicx}       
\graphicspath{{media/}}     
\usepackage[symbol]{footmisc}

\title{LASIGE and UNICAGE
solution to the NASA LitCoin NLP Competition
}

\author{
  Pedro Ruas$^\dagger$, Diana F. Sousa$^\dagger$ \\
  LASIGE \\
  Lisbon, Portugal \\
  \texttt{\{psruas, dfsousa\}@fc.ul.pt} \\
   \And
  André Neves$^\dagger$, Carlos Cruz \\
  Unicage Europe \\
  Lisbon, Portugal\\
  \texttt{\{andre.neves, carlos.cruz\}@unicage.com} \\
  \AND
  Francisco M. Couto \\
  LASIGE \\
  Lisbon, Portugal \\
  \texttt{fcouto@di.fc.ul.pt} \\
}

\begin{document}
\maketitle

\begin{abstract}
Biomedical Natural Language Processing (NLP) tends to become cumbersome for most researchers, frequently due to the amount and heterogeneity of text to be processed. To address this challenge, the industry is continuously developing highly efficient tools and creating more flexible engineering solutions. This work presents the integration between industry data engineering solutions for efficient data processing and academic systems developed for Named Entity Recognition (LasigeUnicage\_NER) and Relation Extraction (BiOnt). Our design reflects an integration of those components with external knowledge in the form of additional training data from other datasets and biomedical ontologies. We used this pipeline in the 2022 LitCoin NLP Challenge, where our team LasigeUnicage was awarded the 7th Prize out of approximately 200 participating teams, reflecting a successful collaboration between the academia (LASIGE) and the industry (Unicage). The software supporting this work is available at \url{https://github.com/lasigeBioTM/Litcoin-Lasige_Unicage}.
\footnotetext[2]{Authors contributed equally to this research.}
\end{abstract}

\keywords{Data Processing \and Named-Entity Recognition \and Relation Extraction \and Knowledge Graphs}

\section{Introduction}

Biomedical data is presented normally in complex, large, and diverse formats. Whether in free-text form or highly specialized knowledge graphs, the data needs to be processed before one can integrate it into predictive pipelines or derive new research hypotheses to target. The data to be processed frequently falls in more than one format and comes in large volumes, which requires efficient data processing \cite{hariri2019uncertainty}. 

Due to the elevated data needs in the industry, there is a need to develop efficient, flexible data engineering solutions to accommodate different formats and volumes. For that, Unicage\footnote{\url{https://unicage.eu/}} offers a set of commands that allow the user to build efficient programs that can be combined in a modular way to build robust, yet flexible, big data processing pipelines. Unicage Europe is a data engineering startup company with a focus on big data processing through the usage of a shell scripting development methodology alongside a set of command-line tools. These commands are format agnostic, meaning they can work with any type of text data. The tools are written in the C programming language and have been geared towards performance, leveraging the OS's memory and resource management capabilities to deliver high-speed processing. The variety of commands is also able to cover gaps that are present in the toolbox of built-in OS utilities. When compared with other big data processing solutions, Unicage tools and methodology are able to fare reasonably well, with cases where it is able to surpass them in terms of speed and efficiency \cite{moreira2018leanbench, https://doi.org/10.48550/arxiv.2212.13647}. 

LASIGE also has vast experience using shell scripting to perform data and text Processing for Health and Life Sciences~\cite{couto-springer2019}.
Lately, LASIGE's NLP academic research focused on two main common tasks, Named-Entity Recognition (NER) and Relation Extraction (RE), by developing systems such as BiOnt \cite{sousa2020biont}. 
These tasks correspond to the 2-phase 2022 LitCoin NLP Challenge\footnote{\url{https://ncats.nih.gov/funding/challenges/litcoin}} in which we participated with our team, LasigeUnicage, creating a successful industry-academia collaboration. Our solution allied the data processing efficiency of Unicage with LASIGE's expertise in NLP. 

The 2022 LitCoin NLP Challenge was a part of the NASA Tournament Lab, hosted by the National Center for Advancing Translational Sciences (NCATS) and the National Library of Medicine (NLM). The competition aimed to create a data-driven technological solution that leverages the vast volumes of biomedical publications published daily to advance the biomedical field by increasing discoverability and formulating new research hypotheses. Specifically, the goal was to extract scientific concepts from scientific articles (Part 1), connect them by generating knowledge assertions, and label them as novel findings or background information (Part 2). 

Our design to target the 2022 LitCoin NLP Challenge relied on two steps: a NER pipeline developed explicitly for the task (Part 1) and the use of the BiOnt system \cite{sousa2020biont} for RE (Part 2). In Part 1, we used the Unicage commands to build a pipeline to gather all the datasets by category and then convert them to the BIO/IBO ("inside-outside-beginning") format required for the NER step. Then, we ensemble six models based on PubMedBERT \cite{PubMedBERT} to recognize the six different types of entities (DiseaseOrPhenotypicFeature, ChemicalEntity, OrganismTaxon, GeneOrGeneProduct, SequenceVariant, CellLine). For Part 2 of the challenge, we used Unicage commands to preprocess the input data and the BiOnt system to perform RE. This system relies not only on the training data itself but can also integrate external knowledge in the form of biomedical ontologies such as ChEBI \cite{degtyarenko2007chebi} and GO \cite{ashburner2000gene} to further improve the RE process. Finally, we resorted to Unicage commands to post-process the output to identify whether the relations were considered novel.     

We state our main contributions described in this paper below:
\begin{itemize}
\item Efficient biomedical NLP Pipeline based on industry data processing tools and academically developed systems for NER and RE. 
\item Application of external data into the biomedical NLP pipeline in the NER and RE stages.
\item Integration of LASIGE's NER and RE systems and Unicage industry data processing solutions.
\end{itemize}

\section{Part 1: Named-Entity Recognition}

In Part 1, given an abstract text the goal was to find/recognize all biomedical entities of types: DiseaseOrPhenotypicFeature, ChemicalEntity, OrganismTaxon, GeneOrGeneProduct, SequenceVariant, or CellLine. For example, given the sentence \textit{Late-onset metachromatic leukodystrophy: molecular pathology in two siblings.}, the goal is to identify the entity \textit{metachromatic leukodystrophy}, a DiseaseOrPhenotypicFeature. 

We started by preprocessing the training datasets containing documents from several Named-Entity Recognition (NER) corpora. Then, we ensemble six trained models (PubMedBERT + linear layer for token classification) to recognize the six different types of entities.

\subsection{Data Processing}

For data processing, we first converted the corpora to the BIO format, merged the different corpora into a single file for each entity type, and generated the final training datasets for each entity type.  Most data was pre-processed using Unicage commands and Shell Scripting. The majority of Unicage commands used were data manipulation commands, such as \textit{self} and \textit{delf}, which are two commands that allow easy manipulation of the data fields on each record of the files, independent of their format, or commands that present an optimization over OS built-in tools, such as \textit{uawk}, which is an optimized version of GNU awk \cite{aho1987awk}. To complement this pipeline, the usage of specific Python libraries, such as bconv\footnote{\url{https://pypi.org/project/bconv/}} and the standoff2conll \footnote{\url{https://github.com/spyysalo/standoff2conll}}, were also used in order to convert the datasets to specific file types so that their manipulation by the Unicage tools could be facilitated.

The datasets used per entity type are the following:

\begin{itemize}

    \item{DiseaseOrPhenotypicFeature}: BC5CDR \cite{bc5cdr}, PGxCorpus \cite{legrand2020}, NCBI Disease \cite{ncbi2014}, Disease Names and Adverse Effects \cite{Gurulingappa2010AnEE}, MedMentions \cite{Mohan2019MedMentionsAL},  and PHAEDRA \cite{Thompson2018AnnotationAD}.

    \item{ChemicalEntity}: Corpora for Chemical Entity Recognition \cite{Kolrik2008ChemicalNT}, CRAFT \cite{bada2012}, BC5CDR \cite{bc5cdr}, CHR \cite{Sahu2019IntersentenceRE}, and PHAEDRA \cite{Thompson2018AnnotationAD}.

    \item{OrganismTaxon}: LINNAEUS \cite{Gerner2010LINNAEUSAS}, CRAFT \cite{bada2012}, Species-800 \cite{Pafilis2013TheSA}, and Cell Finder \cite{cellfinder2012}.

    \item{GeneOrGeneProduct}: BC2GM \cite{Smith2008UvADARED}, JNLPBA \cite{collier2004introduction}, CRAFT \cite{bada2012}, PGxCorpus \cite{legrand2020}, FSU\_PRGE \cite{hahn-etal-2010-proposal}, and Cell Finder \cite{cellfinder2012}.

    \item{CellLine}: JNLPBA \cite{collier2004introduction}, GELLUS \cite{kaewphan2016cell}, CLL \footnote{\url{https://turkunlp.org/Cell-line-recognition/}}, and Cell Finder \cite{cellfinder2012}.

    \item{SequenceVariant}: tmVar \cite{Wei2013tmVarAT}, PGxCorpus \cite{legrand2020}, and SNPPhenA \cite{Bokharaeian2017SNPPhenAAC}.

\end{itemize}

The integration of each training dataset and the competition dataset was done by assigning the relevant tags to the targeted entity. For example, in the training dataset for entities of the type DiseaseOrPhenotypicFeature, the tokens relative to entities in the competition dataset were tagged with \textit{B} and \textit{I}, whereas tokens relative to entities of other types were assigned the tag \textit{O}.

\subsection{Model architecture}

Our approach used PubMedBERT embeddings (originally trained on PubMed articles), jointly with a linear classification layer to classify each token that was fine-tuned in each training dataset.
We defined a post-processing rule for entities of type CHEMICAL with a length of 1. We checked if the character corresponded to a letter since chemical elements can be represented by a single letter (e.g., \textit{C} represents \textit{carbon}). For the remaining entity types, we excluded entities with a length of 1 in the output.

\subsection{Methodology}

For Part 1, our methodology consisted of the following:

\begin{enumerate}
    
    \item{Hyperaparameter optimization}: \texttt{training epochs}, \texttt{learning rate}, \texttt{train batch size}, \texttt{test batch size}. Small versions of the training sets and competition training dataset (in BIO format) were used as test sets. 

    \item{Fine-tuning}: After finding the optimal number of training epochs and learning rates, we combined the competition training dataset with the rest of the training datasets and then trained the model (90\% training, 10\% validation). Training of a distinct model for each of the six competition entity types.

    \item{Prediction}: Sequential application of each model in a given sentence of an abstract present in the competition's test set.
    
\end{enumerate}



\subsection{Implementation}

\begin{itemize}
\item 1 Tesla M10 GPU
\item Training time (excluding hyperparameter optimization) took approximately 15 hours.
\item Prediction time took approximately 8 minutes.
\end{itemize}

\section{Part 2: Relation Extraction}

For Part 2, given the abstract text and the identified entities from Part 1, the goal was to identify relations between the biomedical entities. The relation could be classified in a combination of a first label of Association, Positive Correlation, Negative Correlation, Bind, Cotreatment, Comparison, or Drug Interaction, and a second label of Novel or Not Novel. For example, given the sentence \textit{Midline B3 serotonin nerves in rat medulla are involved in hypotensive effect of methyldopa.} with the identified biomedical entities \textit{serotonin} (ChemicalEntity), \textit{rat} (OrganismTaxon), \textit{hypotensive} (DiseaseOrPhenotypicFeature), and \textit{methyldopa} (ChemicalEntity), the goal is to identify the relation between \textit{serotinin} and \textit{hypontensive} and classify as a Positive Correlation Novel.

Identically to Part 1, we started by preprocessing the datasets provided by the challenge organizers and the biomedical ontologies associated with the entities' identifiers. We assembled two BiOnt \cite{sousa2020biont} models. We used the first model to identify the eight types of relations: Association, Positive Correlation, Negative Correlation, Bind, Cotreatment, Comparison, and Drug Interaction. The second model was to classify the relation even further between Novel and Not Novel. 

\subsection{Data Processing}

Part of the initial data was pre-processed using Unicage commands and Shell Scripting. 

Afterwards, we linked the MESH ontology\footnote{url{https://www.ncbi.nlm.nih.gov/mesh/}} to the entity types DiseaseOrPhenotypicFeature and ChemicalEntity and the NCBITaxon ontology\footnote{url{https://www.ebi.ac.uk/ols/ontologies/ncbitaxon}} to the entity types Species and CellLine.

The competition training set was tokenized using BiOnt, and each entity covered by the ontologies considered above was mapped within the hierarchy of that ontology. 
Finally, the output generated from the model was processed using Unicage commands and Shell Scripting. These scripts used a small set of rules to choose which No/Novel tag to keep for each relation, while at the same time, it generated the final files in the format required by the competition.

\subsection{Model architecture}

Our approach used the BiOnt system, a biomedical RE system built using bidirectional LSTM networks. The BiOnt system incorporates Word2Vec word embeddings \cite{rong2014word2vec} and uses different combinations of input channels to maximize performance, including ontology embeddings.
We used the full provided abstract and considered the multiple relations mentioned within each abstract to have more training cases on the same type of relation.

We are aware of the limitations of our approach, given that the BiOnt system architecture is no longer state-of-the-art for biomedical relation extraction. However, the system's unique approach to external knowledge injection allows us to include each representative ontology within the training pipeline furthering the knowledge about each entity in a candidate relation. 
 
\subsection{Methodology}

For Part 2, our methodology consisted of the following:

\begin{enumerate}
    
    \item{Hyperaparameter optimization}: \texttt{training epochs}, \texttt{learning rate}, \texttt{train batch size}, \texttt{test batch size}, \texttt{max text length}, \texttt{class weights}. Small versions of the training sets and competition training dataset (in BIO format) were used as test sets. 

    \item{Fine-tuning}: After finding the optimal number of training epochs and learning rates, we obtained two trained models, one to predict the different types of relations and the other to predict if they were Novel or not.

    \item{Prediction}: Sequential application of each model in a given abstract present in the competition's dataset.
    
\end{enumerate}

\subsection{Implementtation}

\begin{itemize}
\item 3 Tesla M10 GPU
\item Training time (excluding hyperparameter optimization) took approximately 10 hours.
\item Prediction time took approximately 5 minutes.
\end{itemize}



\section{Evaluation}



For both parts of the challenge (Part 1 and Part 2), we followed the evaluation guidelines provided by the 2022 LitCoin NLP Challenge organizers. The evaluation metric was the average of the Jaccard similarity calculated for each document:

\begin{equation}
    J(O, P) = \frac{|P \cap O|}{|P| + |O| - |P \cup O|}
\end{equation}

\vspace{0.3cm}

In Part 1, $P$ corresponded to the set of predicted mentions and $O$ to the set of correct mentions in a given abstract. A match between two mentions occurred when they had the same type and similar offsets. Our pipeline achieved a score of 0.8423 in this part (calculated from 50.0\% of the test data), which corresponded to 30\% of the final score. The highest-scoring team achieved 0.9067.

In Part 2, $P$ corresponded to the set of predicted relations and $O$ to the set of correct relations in a given abstract. A relation is characterized by a pair of entities, its type (Association, Positive Correlation, Negative Correlation, Bind, Cotreatment, Comparison, and Drug Interaction) and its novelty (No, Novel).

For each correct relation in a given abstract, it is calculated the following intersection score with the predictions:

\begin{equation}
    intersection\_score = 0.25 \times {A} + 0.5 \times {B} + 0.25 \times {C}
\end{equation}

\vspace{0.3cm}

Where $A$, $B$, and $C$ can either have a value of 1 or 0. If a relation in $P$ includes the same pair of entities present in the correct relation, $A$ has a value of 1. If the relation is also of the same type, $B$ has a value of 1. If the relation also has the same novelty, $C$ has a value of 1. This means that the intersection score for a correct relation and the predictions is a value between 0 and 1. The intersection between $O$ and $P$ in a given abstract is calculated by the averaged intersection scores for each correct relation in $O$.

 Our pipeline achieved a score of 0.2124 in this part (calculated from 50.0\% of the test data), which corresponded to 70\% of the final score. The highest-scoring team achieved 0.6279. 

\section{Conclusion}

This work presented the pipeline elaborated to participate in the 2022 LitCoin NLP Challenge by our team, LasigeUnicage, highlighting a successful collaboration between the academia (LASIGE) and the industry (Unicage). Our biomedical NLP pipeline used data engineering pre-processing tools and two systems to perform NER and RE that could incorporate external knowledge. The NER system was explicitly designed to tackle the challenge, whereas, for RE, we used the BiOnt system \cite{sousa2020biont} with minimal modifications. We were awarded the 7th Prize (\$ 5000) in the LitCoin competition out of approximately 200 participating teams. 

In the future, the goal is to improve the NER module by expanding and refining training datasets and exploring different classification layers. As for RE, we could link more external ontological data to boost performance. 
Unicage commands offered advantages, namely in the ease of use and the versatility of the commands, which allowed us to convert and merge the different corpora files to the BIO/IBO format efficiently, along with the post-processing of the results from the Relation Extraction model. We intend to explore how we can further integrate Unicage approaches in NLP tasks and pipelines, with a particular focus on data processing aspects. 

\section*{Acknowledgments}

This work has been supported by FCT through Deep Semantic Tagger (DeST) Project under Grant PTDC/CCIBIO/28685/2017 (\url{http://dest.rd.ciencias.ulisboa.pt/}), in part by LASIGE Research Unit under Grants UIDB/00408/2020 and UIDP/00408/2020, and in part by FCT and FSE through PhD Scholarship under Grant SFRH/BD/145221/2019 and PhD scholarship ref. 2020.05393.BD. 

\bibliographystyle{unsrt}  
\bibliography{main}

\end{document}